\documentclass{article}
\PassOptionsToPackage{numbers,compress}{natbib}

 \usepackage[main, final]{neurips_2026}

\usepackage[utf8]{inputenc} 
\usepackage[T1]{fontenc}    
\usepackage{url}            
\usepackage{booktabs}       
\usepackage{amsfonts}       
\usepackage{nicefrac}       
\usepackage{microtype}      
\usepackage{xcolor}         
\usepackage{pifont}
\usepackage{subcaption}
\usepackage{amsmath}

\usepackage{arydshln}
\usepackage{graphicx}

\usepackage{makecell}
\usepackage{multirow}
\usepackage{colortbl}
\usepackage{array}
\usepackage{wrapfig}

\definecolor{neuripsblue}{rgb}{0.12,0.49,0.85}
\definecolor{lavender}{RGB}{180,140,200}
\definecolor{myPurple}{RGB}{251,237,255}
\usepackage[pagebackref=false,breaklinks,colorlinks,citecolor=neuripsblue]{hyperref}

\def\eg{{\em e.g.}}
\def\ie{{\em i.e.}}

\definecolor{mygray}{gray}{0.9}
\definecolor{highlight}{RGB}{238,250,215}
\definecolor{lightbluegray}{HTML}{8CBBED}

\title{Exploiting Textual Semantics \\for Robust Cross-View Object Correspondence}

\author{%
  Bing Fan \;\;\;\;\; Yunhe Feng \;\;\;\;\; Yan Huang \;\;\;\;\; Heng Fan\thanks{Corresponding author.} \\
  Department of Computer Science and Engineering, University of North Texas\\
}

\begin{document}

\maketitle

\begin{abstract}
    Cross-view object correspondence (CVOC) aims to establish associations of the same target between query and search videos captured from different viewpoints. Existing methods often rely on visual cues for CVOC. Despite recent progress, these methods struggle in complicated scenarios with substantial viewpoint changes and appearance variations, due to lacking sufficient target information. Addressing this, we introduce \textbf{\emph{TeSCo}}, a novel framework that exploits rich \textbf{Te}xtual \textbf{S}emantics of an object, in addition to its visual cues, for cross-view object \textbf{Co}rrespondence. Our \emph{key insight} is, textual description of a target, by capturing diverse attributes such as category, texture, and color, provides \emph{viewpoint-} and \emph{appearance-invariant} semantics, which are complementary to visual cues and can hence enhance correspondence robustness in complicated scenes. Inspired by this, TeSCo first generates a textual expression of the target from the given query view using a vision-language model, and then extracts textual feature to enhance visual features for localization in the search view. To realize this, we introduce a simple yet effective \emph{text-conditioned cross-modal fusion} (TCF) module that incorporates the textual feature into visual features with text-guided modulation, producing more robust multimodal target representation for object correspondence. Since not all textual cues are equally compatible with the target in current search-view frame, we propose an \emph{iterative textual feature refinement} (ITFR) module that progressively adjusts textual feature using search-view information before the TCF module, enabling search-view-aware textual semantics to be fused into visual features and leading to better performance. In extensive experiments on Ego-Exo4D and HANDAL-X, TeSCo demonstrates state-of-the-art results and largely outperforms other models, validating its efficacy. Our code is released at \url{https://github.com/VAIL-Fan/TeSCo}. 
  
\end{abstract}

\section{Introduction}

Cross-view object correspondence (CVOC) aims to establish associations of the same target across different viewpoints, \eg, ego-centric and exo-centric views~\cite{grauman2024ego}. In specific, given a query-view video with frame-wise target mask annotations, the goal is to segment the same object in the corresponding frames of a search-view video. As an important problem in multi-view video understanding, CVOC enables many crucial applications, such as robot learning, augmented reality, human-object interaction understanding, thereby attracting growing interest in recent years (\eg,~\cite{baade2025self,fu2025objectrelator,hu2025robust,liao2025domr,gao2026vggt,yan2026learning,pan2026v}).

Current models predominantly rely on visual cues for CVOC. One line of work (\eg,~\cite{baade2025self,liao2025domr,gao2026vggt,yan2026learning,pan2026v}) focuses on directly learning frame-level cross-view visual correspondence through extracting and matching visual features using image pairs across query and search views, while another line of work (\eg,~\cite{hu2025robust}) leverages both cross-view image pairs and temporal modeling of visual features within each viewpoint to enhance correspondence. Despite their progress, these approaches often struggle in challenging cross-view scenarios with substantial viewpoint and appearance changes, because visual representations learned by existing methods are \emph{inherently sensitive} to such visual variations, making it difficult to maintain consistent target representations across views when relying on only visual cues and resulting in degraded performance. This limitation motivates the search for complementary target information that is inherently more robust to viewpoint and appearance changes for CVOC.

In this paper, we leverage textual cues for CVOC. Our \emph{key insight} is that, the textual description of an object (\eg, ``\emph{a red basketball with black line, white stripe, and leather surface}'') often captures rich high-level attributes, such as \emph{category}, \emph{color}, \emph{material}, and \emph{texture}. While the visual appearance of a target may vary drastically across views and over time in complicated scenarios, its attributes remain \emph{consistent}, providing partially more stable semantic cues that are complementary to visual information and can therefore enhance the robustness of cross-view object correspondence.

\begin{figure}[!t]
    \centering
    \includegraphics[width=1\linewidth]{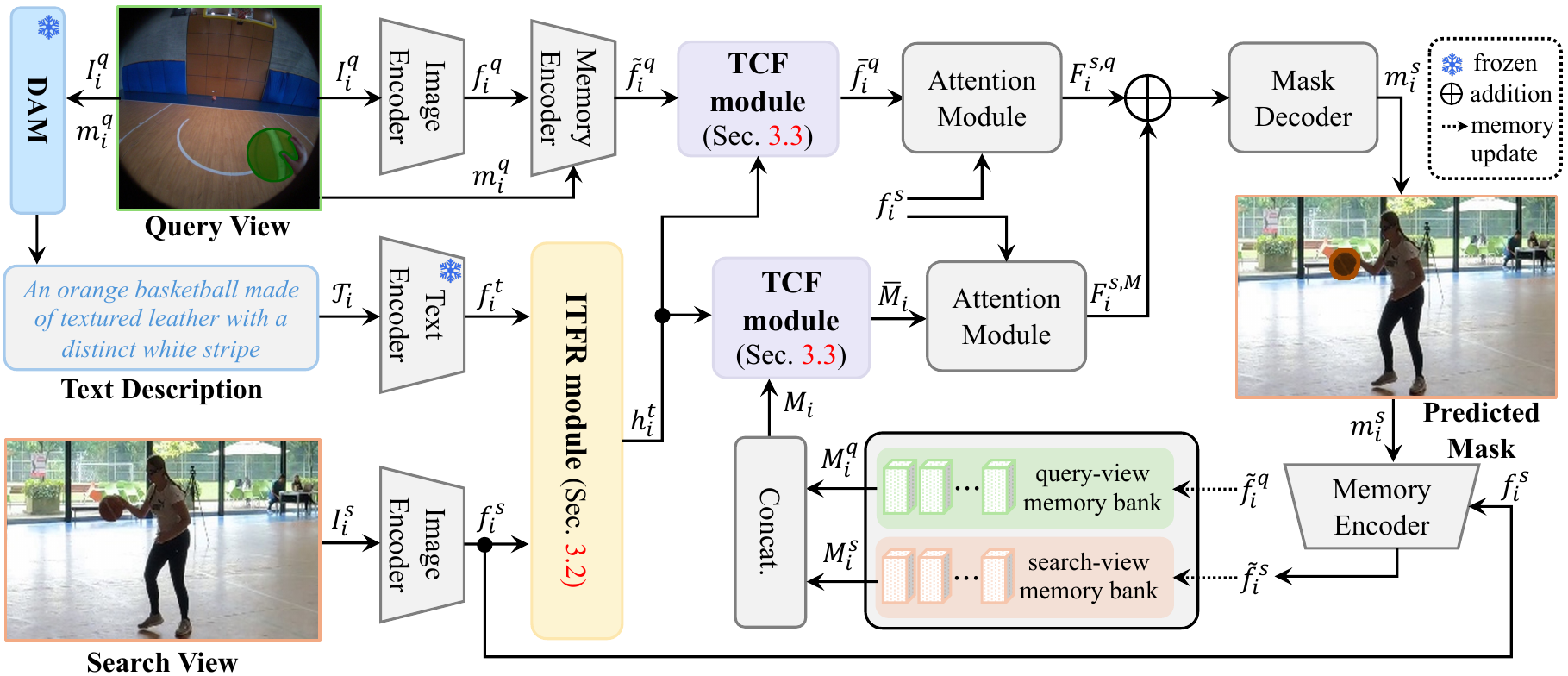}
    \caption{Overview of the proposed TeSCo that exploits rich textual information for robust cross-view correspondence through iterative textual feature refinement (ITFR) and text-conditioned cross-modality fusion (TCF) modules. \emph{Best viewed in pdf for all figures throughout the paper}.}
    \label{framework}
\end{figure}

Motivated by the above, we introduce \textbf{\emph{TeSCo}}, a novel framework that exploits rich \textbf{Te}xtual \textbf{S}emantics of the target, in addition to its visual cues, for cross-view object \textbf{Co}rrespondence. Specifically, TeSCo, is built on top of the popular SAM 2 model~\cite{RaviGHHR0KRRGMP25}. Its core idea is to leverage the textual information of the object to enhance spatial view and temporal memory features, and employ the resulted multimodal representation, which is more robust against the view and appearance variations, for cross-view object correspondence. Since the textual description is not provided in the standard CVOC task, we first generate a detailed textual description of the target from the given query-view frame using an off-the-shelf vision-language model, and then extract its textual feature to enhance visual features for correspondence. To realize this, we introduce a simple yet effective \emph{text-conditioned cross-modal fusion} (TCF) module, which injects the textual information into visual features with text-guided modulation. TCF enables selective enhancement of target-relevant cues in fusion using text, resulting in more robust multimodal target representation for cross-view object correspondence. In addition, considering that not all textual semantics are equally compatible with the current search frame, we propose an \emph{iterative textual feature refinement} (ITFR) module in TeSCo, which progressively adjusts the textual feature using information in search view before cross-modal fusion in the TCF module, producing search-view-aware textual semantics for more effective feature fusion. Through the ITFR and TCF modules, our TeSCo is able to exploit rich textual information of the target to enhance visual features for more robust CVOC. Fig.~\ref{framework} illustrates the framework of TeSCo.

We note that an early approach~\cite{fu2025objectrelator} has also explored the use of text for cross-view correspondence. Although TeSCo shares the general idea of applying text, it substantially differs from~\cite{fu2025objectrelator}. \emph{First}, the text in~\cite{fu2025objectrelator}, containing only object name, is primarily utilized as part of prompt to instruct the large language model (LLM) for segmentation, whereas TeSCo adopts richer textual semantics, including category, color, texture, \emph{etc.}, as feature-level guidance to explicitly enhance spatial and temporal visual features for correspondence. \emph{Second}, different from~\cite{fu2025objectrelator} in applying cross-attention to merge text and visual embeddings, TeSCo introduces a dedicated TCF module that further refines feature fusion using text, enabling further enhancement of target-relevant cues for more robust multimodal representation. \emph{Third}, unlike~\cite{fu2025objectrelator} directly using the text feature, we introduce an ITFR module to adjust the textual feature using search-view information, leading to search-view-aware textual semantics for better performance. Compared with~\cite{fu2025objectrelator}, our TeSCo more effectively exploits textual semantics from a different perspective and achieves more robust performance, as demonstrated in later experiments.

To validate the effectiveness of TeSCo, we conduct extensive experiments on two benchmarks, including Ego-Exo4D~\cite{grauman2024ego} and HANDAL-X~\cite{guo2023handal}, and compare it with existing models. Our evaluation shows that TeSCo achieves new state-of-the-art results, with Intersection over Union (IoU) scores of \textbf{60.4\%} in the ego$\rightarrow$exo setting and \textbf{70.3\%} in the exo$\rightarrow$ego setting on the Ego-Exo4D and \textbf{90.0\%} on the HANDAL-X, which largely surpass current methods and its baseline SAM 2, showing its efficacy.

In summary, our \textbf{\emph{contributions}} are as follows: \ding{171} we propose a framework, named TeSCo, that exploits textual cues as semantic guidance to enhance CVOC and achieves state-of-the-art performance in extensive experiments on Ego-Exo4D and HANDAL-X; \ding{170} we present the TCF module, which injects textual semantics into visual features with text-guided modulation to enhance target-relevant cues; \ding{168} we further design an ITFR module, which adapts textual features with search-view information before fusion, producing search-view-aware textual semantics for more effective feature fusion.



\section{Related Work} 

\textbf{Cross-View Object Correspondence.} Cross-view object correspondence (CVOC) aims to associate the same target between different viewpoints. Owing to its important applications, CVOC has recently received extensive attentions with many methods proposed. The work of~\cite{baade2025self} designs a self-supervised framework for cross-view correspondence with predictive cycle consistency. The approaches in~\cite{liao2025domr,mur2025mama} propose matching masks between different views to achieve object correspondence. The work in~\cite{yan2026learning} presents cycle-consistency training to learn view-invariant visual representations for cross-view correspondence. The work of~\cite{fu2025objectrelator} utilizes a large language model with visual and text prompts for CVOC. The methods in~\cite{hu2025robust,pan2026v} explores SAM 2~\cite{RaviGHHR0KRRGMP25} for cross-view segmentation. The model of~\cite{gao2026vggt} applies geometry and appearance features in~\cite{wang2025vggt} to learn robust representation for CVOC. 

In this work, we introduce TeSCo that exploits rich textual semantics for cross-view correspondence. TeSCo is built upon SAM 2 by leveraging its excellent segmentation capacity. However, different from existing methods~\cite{hu2025robust,pan2026v} that explores visual cues only within the SAM 2 framework, TeSCo leverages both visual and textual cues for cross-view correspondence, leading to more robust performance. In addition, despite sharing a similar spirit, TeSCo is significantly different from~\cite{fu2025objectrelator} in how text is used and adapted to enhance CVOC as discussed earlier.

\vspace{0.3em}
\textbf{Visual Correspondence.} Visual correspondence aims to establish associations of the same visual element across different frames or viewpoints, and is a crucial computer vision problem. Depending on the granularity of the association, visual correspondence can be divided into two groups, including point-level correspondence such as optical flow~\cite{lu2024promotion,jin2025flovd} and point tracking~\cite{karaev2024cotracker,karaev2025cotracker3}, which focuses on fine-grained matching of pixels, and object-level correspondence such as pedestrian search~\cite{xiao2017joint}, video segmentation~\cite{RaviGHHR0KRRGMP25,cheng2022xmem} and tracking~\cite{lin2024tracking}, and visual query localization~\cite{khosla2025relocate,fan2025prvql}, which aims to associate the entire target. TeSCo falls in the object-level correspondence category. However, different from existing object correspondence methods, TeSCo focuses on associating the target in videos captured by different viewpoints, which is challenging due to extreme variations in appearance and viewpoint.

\vspace{0.3em}
\textbf{Text in Visual Localization.} Textual description provides rich semantics of an object, and has been widely applied in various visual localization tasks, such as visual grounding~\cite{zhu2022seqtr,yang2022improving,liu2024grounding}, spatial-temporal video grounding~\cite{yang2022tubedetr,gu2024context,gu2025knowing}, referring multi-object tracking~\cite{wu2023referring,li2025language}, referring expression segmentation~\cite{liu2023gres,hu2023beyond,chng2024mask}, and vision-language tracking~\cite{guo2022divert,feng2024memvlt}, in which the text is offered to localize the target of interest in the image or video. Different from the above tasks, in this paper we focus on exploiting the textual information for improving cross-view object correspondence.

\section{The Proposed Method}

\textbf{Task Definition.} Cross-view object correspondence aims to associate the same target across different viewpoints. Formally, given a query-view (\eg, ego-centric) sequence $\{I_i^{q}\}_{i=1}^{N}$ with $N$ frames and masks $\{m_i^{q}\}_{i=1}^{N}$ of a target of interest in each frame, the goal is to segment the object masks $\{m_i^{s}\}_{i=1}^{N}$ of the same target in each frame of a synchronized search-view (\eg, exo-centric) video $\{I_i^{s}\}_{i=1}^{N}$.


\vspace{0.2em}
\textbf{Overview.} In this paper, we introduce TeSCo, which leverages textual information to enhance target representation for improving CVOC. TeSCo is built upon SAM 2~\cite{RaviGHHR0KRRGMP25} for its excellent segmentation capacity. Specifically, as displayed in Fig.~\ref{framework}, TeSCo first generates a detailed textual description of the target from the given query-view frame, and then applies it to enhance the visual features in SAM 2 for target segmentation ($\S$\ref{tesco}). To realize this, two simple yet effective modules, including ITFR ($\S$\ref{itm}) and TCF ($\S$\ref{tcf}), are introduced, with ITFR to refine the textual feature using information in current search-view frame and TCF to inject the refined textual information into visual features with text-guided modulation for learning multimodal features for segmentation. 

\subsection{Our TeSCo Framework}
\label{tesco}

In the following, we detail TeSCo by describing how textual information is leveraged to segment the target object from the $i^{\text{th}}$ frame $I_i^{s}$ in the search-view video.

\vspace{0.2em}
\textbf{Textual Description Generation.} Since the text is \emph{not} provided in the standard CVOC task, we first utilize a pre-trained vision-language model DAM~\cite{lian2025describe} to generate a detailed textual description of the target from the given query-view frame. The reason for employing DAM is its excellent capability for detailed image region captioning. Specifically, given the $i^{\text{th}}$ frame $I_i^q$ in the query-view video with the target mask $m_i^q$, the textual description of the target is generated as follows, 
\begin{equation}
    \mathcal{T}_i = \mathtt{DAM}(I_i^q,m_i^q)
\end{equation}
where $\mathcal{T}_i$ is the generated text in frame $I_i^q$, which covers category, color, texture, \emph{etc}. Due to limited space, please refer to the \emph{\textbf{appendix}} for examples of $\mathcal{T}_i$. It is worth noting that, although $\mathcal{T}_i$ is generated from the query view, its encoded textual semantics capture attributes that are relatively stable across views, and can thus assist target segmentation in the search view. 

In this work, we generate the textual description $\mathcal{T}_i$ from each query-view frame $I_i^q$. This is because the target may reveal different visible attributes at different moments, and the frame-specific description can provide more suitable semantics for localization in the corresponding search-view frame $I_i^s$. 

\vspace{0.2em}
\textbf{Exploiting Textual Semantics for CVOC.} After generating $\mathcal{T}_i$, TeSCo exploits its textual semantics for improving CVOC. Specifically, TeSCo is built upon SAM 2~\cite{RaviGHHR0KRRGMP25}, and applies text to enhance the query-view and temporal memory features for target segmentation in $I_i^s$. 

Given a pair of $I_i^q$ and $I_i^s$, we first extract their features $f_i^q=\Phi(I_i^q)$ and $f_i^s=\Phi(I_i^s)$, where $\Phi(\cdot)$ is the image encoder~\cite{ryali2023hiera} in SAM 2. After that, $f_i^q$, together with the object mask $m_i^q$, are applied to generate the object-aware query-view feature $\tilde{f}_i^q=\texttt{MemEnc}(f_i^q,m_i^q)$, where $\texttt{MemEnc}(\cdot,\cdot)$ denotes the memory encoder of SAM 2. Besides the query-view feature $\tilde{f}_i^q$, the temporal memory feature, formed by memory banks from both query and search views, is exploited in TeSCo. Let $M_{i}^q$ and $M_{i}^s$ denote query-view and search-view memory banks in frame $i$. For $i=1$, both $M_{i}^q$ and $M_{i}^s$ are empty. Then, the temporal memory feature $M_i=\texttt{Concat}(M_{i}^q,M_{i}^s)$ is obtained by concatenating $M_{i}^q$ and $M_{i}^s$.

Different from~\cite{hu2025robust} directly using query-view and memory features for CVOC, we apply textual cues to enhance visual features and then use the resulted more robust multimodal features for segmentation. Specifically, given the textual description $\mathcal{T}_i$, we first extract its feature $f_i^t=\Psi(\mathcal{T}_i)$, where the text encoder $\Psi(\cdot)$ adopts RoBERTa~\cite{liu2019roberta}. Considering that not all textual cues in $\mathcal{T}_i$ are equally important in distinguishing the target in $I_i^s$ due to the gap between different views, we apply ITFR (as described later in Sec.~\ref{itm}) to refine $f_i^t$ using information in the search-view frame, as follows,
\begin{equation}
    h_i^t = \mathtt{ITFR}(f_i^t,f_i^s)
\end{equation}
where $h_i^t$ denotes the refined text feature, which focuses more on textual semantics that are compatible with the search-view frame $I_i^s$ for segmentation. Afterwards, we inject the textual cues into $\tilde{f}_i^q$ and $M_i$ through the TCF module (as explained later in Sec.~\ref{tcf}), as follows,
\begin{equation}
    \Bar{f}_i^q = \mathtt{TCF}(\tilde{f}_i^q,h_i^t) \;\;\;\;\;\;\;\;
    \Bar{M}_i = \mathtt{TCF}(M_i,h_i^t)
\end{equation}
where $\Bar{f}_i^q$ and $\Bar{M}_i$ are enhanced multimodal query-view and memory features. Compared to $\tilde{f}_i^q$ and $M_i$, they are more robust against view and appearance changes, benefiting cross-view correspondence.

After this, we fuse $\Bar{f}_i^q$ and $\Bar{M}_i$ into the search-view feature $f_i^s$ via attention module, as follows,
\begin{equation}
    F_i^{s,q}=\mathtt{MemAtt}(f_i^s,\Bar{f}_i^q) \;\;\;\;\;\;\;\; F_i^{s,M}=\mathtt{MemAtt}(f_i^s,\Bar{M}_i)
\end{equation}
where $\mathtt{MemAtt}(\cdot,\cdot)$ denotes the memory attention module in SAM 2. The two outputs are combined and fed into the mask decoder to predict the target mask in the search-view frame, as follows,
\begin{equation}
    m_i^s=\mathtt{MaskDec}(F_i^{s,q} + F_i^{s,M})
\end{equation}
where $m_i^s$ denotes the predicted target mask in $I_i^s$. After obtaining $m_i^s$, we generate the object-aware search-view feature $\tilde{f}_i^s$ via $\tilde{f}_i^s=\mathtt{MemEnc}(f_i^s,m_i^s)$, and apply $\tilde{f}_i^s$ and $\tilde{f}_i^q$ to update the memory banks $M_{i}^s$ and $M_{i}^q$ to $M_{i+1}^s$ and $M_{i+1}^q$ with the first-in-first-out strategy for the $(i+1)^{\text{th}}$ frame.

\subsection{Iterative Textual Feature Refinement (ITFR)}
\label{itm}

Accurate textual cues are important for robust target segmentation in the search-view frame. Although the text captures semantic attributes that are relatively stable across different viewpoints, some cues may become \emph{less} applicable to the search view because of the cross-view appearance variations. For example, certain attributes such as texture in the query-view description may become ambiguous or even invisible in the search view. To deal with this issue, we propose an iterative textual feature refinement (ITFR) module, which progressively adjusts the textual feature using search-view information, enabling better adaption to current search-view context before being fused with visual features.

\setlength{\columnsep}{10pt}%
\setlength\intextsep{0pt}
\begin{wrapfigure}{r}{0.5\textwidth}
\centering
\includegraphics[width=0.5\textwidth]{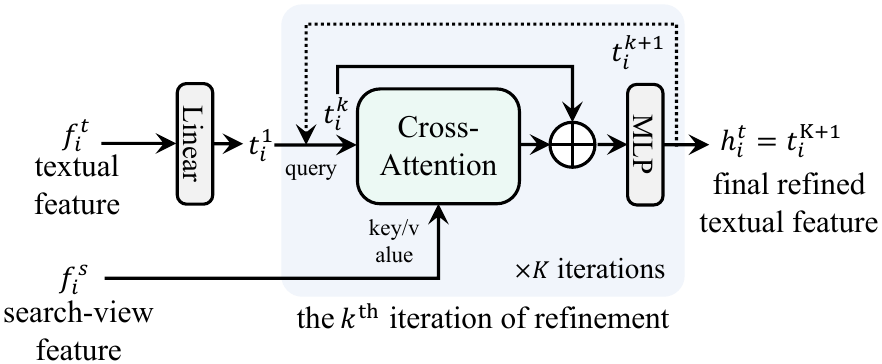}
\caption{Illustration of the ITFR module.}
\label{fig:itf}
\end{wrapfigure}
Specifically, ITFR comprises $K$ iterations of textual feature refinement. In each iteration, the refined textual feature from last stage is interacted with search-view feature for further adjustment. Concretely, let $t_i^k$ represent the input textual feature in iteration $k$ ($1\le k \le K$). In the first iteration ($k=1$), $t_i^1$ is initialized via $t_i^1=\mathtt{Linear}(f_i^t)$, where $\mathtt{Linear}(\cdot)$ is a linear projection layer that maps the channel dimension of the textual feature to match that of $f_i^s$. To adapt the textual representation to the current search-view context, in each iteration, we first adopt cross-attention~\cite{vaswani2017attention} to inject search-view information into the textual feature, and then employ a residual connection to preserve the original textual semantics while incorporating the search-view-guided refinement, followed by a multilayer perceptron (MLP) module. Mathematically, this process can be expressed as follows,
\begin{equation}
    t_i^{k+1} = \mathtt{MLP}(\mathtt{CA}(t_i^k,f_i^s) + t_i^k) \;\;\;\;\;\;\;\; k=1,2,\cdots,K
\end{equation}
where $\mathtt{CA}(z,u)$ denotes cross-attention with $z$ and $u$ the query and key/value, respectively. After $K$ iterations, the final refined textual feature $h_i^t$ is obtained as $h_i^t=t_i^{K+1}$, which will be fed into the TCF module for multimodal fusion. Fig.~\ref{fig:itf} illustrates the ITFR for textual feature refinement.

It is worth \textbf{\emph{noting}} that, while ITFR is implemented with a simple cross-attention-based structure, it effectively adapts the query-view textual representation to the current search-view context. Through ITFR, the textual feature progressively incorporates search-view visual information, suppressing less irrelevant attributes caused by cross-view variations and enhancing textual semantics that are more compatible with the current search-view frame.

\subsection{Text-Conditioned Cross-Modality Fusion (TCF)}
\label{tcf}

\setlength{\columnsep}{10pt}%
\setlength\intextsep{0pt}
\begin{wrapfigure}{r}{0.5\textwidth}
\centering
\includegraphics[width=0.5\textwidth]{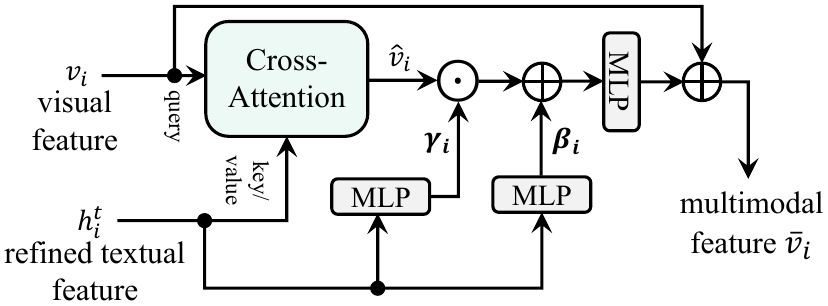}
\caption{Illustration of the TCF module.}
\label{fig:tcf}
\end{wrapfigure}
To improve cross-view object segmentation, we use textual semantics to enhance target representation. Specifically, after obtaining the refined textual feature, we fuse it into the query-view feature and the temporal memory feature, and employ the resulting multimodal features for target segmentation. To realize this, we introduce a simple yet effective text-conditioned cross-modality fusion (TCF) module, which injects the textual feature into the visual representation with text-guided modulation, enabling better enhancement of target-relevant cues in fusion. Fig.~\ref{fig:tcf} illustrates the architecture of TCF.

Concretely, given the refined textual feature $h_i^t$ and the visual feature $v_i$, where $v_i$ represents either the query-view feature $\tilde{f}_i^q$ or the memory feature $M_i$, we first perform cross-attention to integrate the textual information into the visual feature as follows,
\begin{equation}
    \widehat{v}_i = \mathtt{CA}(v_i,h_i^t)
\end{equation}
where $\widehat{v}_i$ denotes the multimodal target representation that contains both visual and textual information. To further enhance target-relevant cues in $\widehat{v}_i$, we leverage the textual feature $h_i^t$ as a guidance to modulate the multimodal target feature. Specifically, inspired by conditional normalization~\cite{perez2018film,de2017modulating}, we generate two text-conditioned vector parameters, $\boldsymbol{\gamma}_i$ and $\boldsymbol{\beta}_i$, to modulate $\widehat{v}_i$, with MLPs via 
\begin{equation}
    \boldsymbol{\gamma}_i = \mathtt{MLP}(h_i^t) \;\;\;\;\;\;\;\; \boldsymbol{\beta}_i = \mathtt{MLP}(h_i^t)
\end{equation}
Here, $\boldsymbol{\gamma}_i$ is a channel-wise reweighting parameter that selectively emphasizes target-relevant cues in $\bar{v}_i$, while $\boldsymbol{\beta}_i$ acts as a shift parameter that further adapts the feature distribution according to textual semantics. With $\boldsymbol{\gamma}_i$ and $\boldsymbol{\beta}_i$, the final multimodal representation is then obtained via modulation with a residual connection, as follows,
\begin{equation}\label{mod}
    \bar{v}_i = \mathtt{MLP}(\widehat{v}_i \odot \boldsymbol{\gamma}_i + \boldsymbol{\beta}_i) + v_i
\end{equation}
where $\odot$ denotes element-wise multiplication, and $\bar{v}_i$ the final multimodal feature, which corresponds $\bar{f}_i^q$ when TCF is applied to $\tilde{f}_i^q$, or $\bar{M}_i$ when TCF is applied to $M_i$. It is worth \emph{\textbf{noting}} that, in practice, the vector parameters $\boldsymbol{\gamma}_i$ and $\boldsymbol{\beta}_i$ are broadcast along the spatial dimension in Eq. (\ref{mod}) to match the shape of $\widehat{v}_i$. Through this text-guided modulation, the resulting multimodal representation can better focus on target-relevant cues for cross-view segmentation. 

\begin{figure}[!t]
    \centering
    \includegraphics[width=1\linewidth]{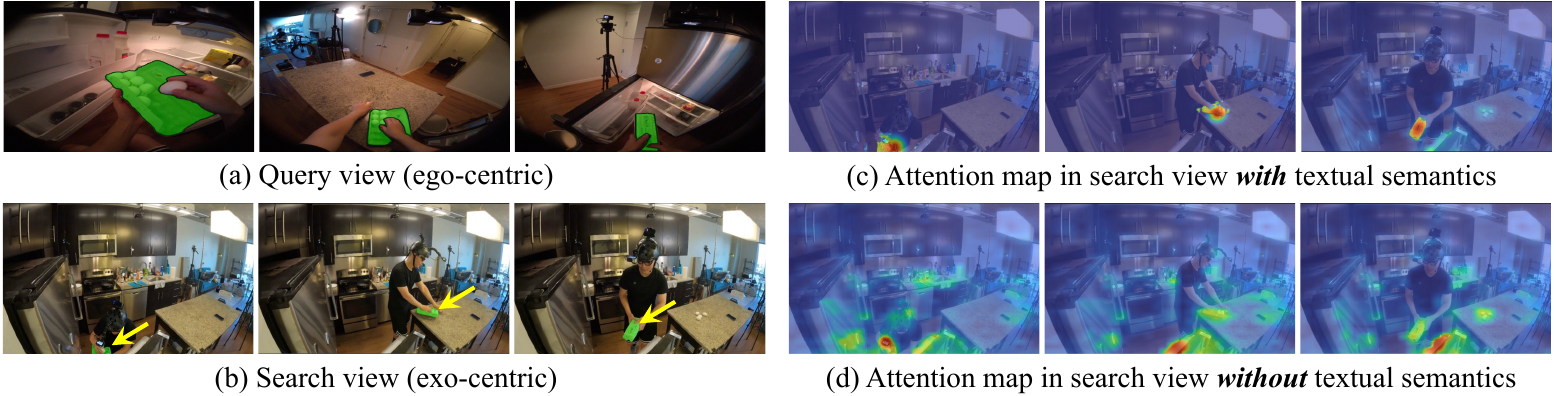}
    \caption{Image (a): query-view frames; image (b): search-view frames; image (c): attention maps in search view with textual semantics; and (d) attention maps in search view without textual semantics. The visualized attention maps is obtained by aggregating the attention maps from the two attention modules in TeSCo. We can observe that, the attention map with textual semantics can more accurately focus on the target region, evidencing the effectiveness of textual information.}
    \label{attmap}
\end{figure}

Compared with the visual features $\tilde{f}_i^q$ and $M_i$, the multimodal features $\bar{f}_i^q$ and $\bar{M}_i$, by integrating textual semantics, which are view- and appearance invariant, are able to better preserve discriminative target cues under cross-view appearance variations, thereby improving the robustness of cross-view object correspondence. As demonstrated in Fig.~\ref{attmap}, with textual semantics incorporated through TCF, the resulting attention map in the subsequent attention modules focuses more accurately on the target region, validating the effectiveness of textual semantics.

\subsection{Optimization}

Given a query-view video $\{I_i^{q}\}_{i=1}^{N}$ with object masks $\{m_i^{q}\}_{i=1}^{N}$ and a search-view sequence $\{I_i^{s}\}_{i=1}^{N}$, our TeSCo predicts the target mask $m_i^{s}$ in each search-view frame $I_i^{s}$. During training, we follow SAM 2~\cite{RaviGHHR0KRRGMP25}, and employ multiple losses, including dice coefficient loss, focal mask loss, IoU loss, and binary cross-entropy loss to supervise the predicted mask $m_i^{s}$ in each search-view frame. The final training objective is obtained by summing the losses over all predicted search-view masks. Because of the limitation of space, please kindly refer to SAM 2~\cite{RaviGHHR0KRRGMP25} for more details and implementations of the loss function for training.

\section{Experiments}

\textbf{Implementation.} TeSCo is built upon SAM 2~\cite{RaviGHHR0KRRGMP25}, and introduces two dedicated modules, TCF and ITFR, to exploit the textual cues for cross-view correspondence. It is implemented using PyTorch~\cite{paszke2019pytorch} and trained with 4 Nvidia A6000 GPUs. Following~\cite{RaviGHHR0KRRGMP25}, we utilize the HieraB+~\cite{ryali2023hiera} as the visual encoder to extract visual features. For the text, we apply RoBERTa~\cite{liu2019roberta} to extract its feature. Our TeSCo is end-to-end trained for 60 epoches with a batchsize of 12 (3 per GPU) using the AdamW optimizer~\cite{LoshchilovH19}. The learning rates for the image encoder and our new modules/decoder are $6\times10^{-6}$ and $10^{-5}$, respectively. During training and inference, all video frames are resized to $480\times480$ because of GPU memory constraints. The number of iteration $K$ in ITFR is empirically set to 2. The loss function for training is kept the same as in~\cite{RaviGHHR0KRRGMP25}. Code and model will be released.

\vspace{0.2em}
\textbf{Datasets.} We use Ego-Exo4D~\cite{grauman2024ego} and HANDAL-X~\cite{guo2023handal} for experiments. Ego-Exo4D~\cite{grauman2024ego} contains 1,335 videos takes and around 1.8M annotated object masks (742K in the ego-centric view and 1.1M in the exo-centric view), covering diverse real-world scenarios such as basketball, bike Repair, cooking, health, and music. Following previous works~\cite{hu2025robust,yan2026learning,pan2026v}, we employ the official training/test split in Ego-Exo4D and adopt two settings, Ego$\rightarrow$Exo (Ego-centric view as query) and Exo$\rightarrow$Ego (Exo-centric view as query), for experiments. HANDAL-X~\cite{guo2023handal}, introduced in~\cite{fu2025objectrelator}, consists of real-world manipulable objects with 6-DoF pose labels and is used to further assess the generalization ability of different models. It contains 44,102 training pairs and 14,074 test pairs for experiments.

\vspace{0.2em}
\textbf{Evaluation Metrics.} Following~\cite{pan2026v}, we use \emph{Intersection over Union} (IoU), \emph{Location Error} (Loc.E), and \emph{Contour Accuracy}
(Cont.A), provided in~\cite{grauman2024ego}, as the evaluation metrics on Ego-Exo4D. According to Ego-Exo4D Correspondence Challenge~\cite{grauman2024ego}, the primary evaluation metric is the mean Intersection-over-Union (mIoU) of both Ego$\rightarrow$Exo and Exo$\rightarrow$Ego tasks. Please refer to~\cite{grauman2024ego} for the details of these metrics. On HANDAL-X, following prior works~\cite{fu2025objectrelator,yan2026learning}, we utilize IoU for evaluation.


\subsection{State-of-the-art Comparison}

\textbf{Results on Ego-Exo4D.} To verify the effectiveness of TeSCo, we evaluate it on Ego-Exo4D with comparison against existing state-of-the-art methods and its baseline. The baseline is implemented by removing the ITFR and TCF modules form TeSCo. Due to limited space, we display the architecture of the baseline in the \emph{\textbf{appendix}}.

\renewcommand{\arraystretch}{1.05}
\begin{table}[!t]
  \centering
  \caption{Results of Ego$\rightarrow$Exo and Exo$\rightarrow$Ego on the Ego-Exo4D test set. The best and the second-best results are highlighted in \textbf{bold} and \underline{underlined} in all state-of-the-art comparison tables.}
  \vspace{5pt}
  \resizebox{0.98\textwidth}{!}{
    \begin{tabular}{rccccccc}
    \Xhline{1.2pt}
          & \multicolumn{3}{c}{\textbf{Ego$\rightarrow$Exo}} & \multicolumn{3}{c}{\textbf{Exo$\rightarrow$Ego}} & \multirow{2}{*}[-0.8ex]{\makecell[c]{\textbf{mIoU}\\ \textbf{($\uparrow$)}}} \\
    \cmidrule(lr){2-4} \cmidrule(lr){5-7}
    \textbf{Method} & \textbf{IoU ($\uparrow$)}   & \textbf{Cont.A ($\uparrow$)} & \textbf{Loc.E ($\downarrow$)} & \textbf{IoU ($\uparrow$)} & \textbf{Cont.A ($\uparrow$)} & \textbf{Loc.E ($\downarrow$)} &  \\
    \hline\hline
    XSegTx\textcolor{gray}{\scriptsize{[CVPR'24]}}~\cite{grauman2024ego} & 18.90 & 0.386 & 0.070 & 27.10 & 0.358 & 0.104 & 23.00 \\
    XMem\textcolor{gray}{\scriptsize{[CVPR'24]}}~\cite{grauman2024ego}  & 19.30 & 0.262 & 0.151 & 16.60 & 0.240 & 0.160 & 17.95 \\
    XMem+XSegTx\textcolor{gray}{\scriptsize{[CVPR'24]}}~\cite{grauman2024ego} & 34.90 & 0.559 & 0.038 & 25.00 & 0.237 & 0.117 & 29.95 \\
    PCC\textcolor{gray}{\scriptsize{[CVPR'25]}}~\cite{baade2025self}   & 39.01 & 0.600 & 0.033 & 47.06 & 0.590 & 0.054 & 43.04 \\
    ObjectRelator\textcolor{gray}{\scriptsize{[ICCV'25]}}~\cite{fu2025objectrelator} & 35.30 & 0.540 & 0.036 & 40.30 & 0.500 & 0.068 & 37.80 \\
    O-MaMa (k-NN)\textcolor{gray}{\scriptsize{[ICCV'25]}}~\cite{mur2025mama} & 31.90 & 0.414 & 0.195 & 30.90 & 0.373 & 0.127 & 31.40 \\
    O-MaMa\textcolor{gray}{\scriptsize{[ICCV'25]}}~\cite{mur2025mama} & 42.60 & 0.590 & 0.033 & 44.10 & 0.524 & 0.082 & 43.35 \\
    CCMP\textcolor{gray}{\scriptsize{[CVPR'26]}}~\cite{yan2026learning}  & 41.95 & 0.669 & 0.038 & 47.18 & 0.591 & 0.081 & 44.57 \\
    V$^{2}$-SAM\textcolor{gray}{\scriptsize{[CVPR'26]}}~\cite{pan2026v}  & 46.30 & 0.616 & 0.056 & 49.60 & 0.575 & 0.078 & 47.95 \\
    LM-EEC\textcolor{gray}{\scriptsize{[NeurIPS'25]}}~\cite{hu2025robust} & \underline{54.98} & \underline{0.778} & \underline{0.017} & \underline{65.77} & \underline{0.774} & \underline{0.031} & \underline{60.38} \\
    \arrayrulecolor{gray!50}
    \hline
    \arrayrulecolor{black}
    Baseline & 52.70 & 0.719 & 0.027 & 58.30 & 0.664 & 0.043 & 55.50 \\
    \rowcolor[HTML]{E8FAF9}
    TeSCo (ours) & \textbf{60.40} & \textbf{0.847} & \textbf{0.011} & \textbf{70.30} & \textbf{0.808} & \textbf{0.023} & \textbf{65.35} \\
    \Xhline{1.2pt}
    \end{tabular}}
  \label{tab:egoexo_res}%
\end{table}%

Tab.~\ref{tab:egoexo_res} presents the results for Ego$\rightarrow$Exo and Exo$\rightarrow$Ego. As shown, TeSCo consistently achieves the best performance across all metrics in both settings. Specifically, it achieves IoU scores of 60.40\% for Ego$\rightarrow$Exo and 70.30\% for Exo$\rightarrow$Ego, with an overall mIoU of 65.35\%, significantly outperforming the second best method, LM-EEC~\cite{hu2025robust}, which achieves 54.98\% and 65.77\% IoU scores and a mean IoU of 60.38\%. Notably, although LM-EEC is also built upon SAM 2~\cite{RaviGHHR0KRRGMP25}, it primarily relies on long-term visual cues. The superior performance of TeSCo over LM-EEC highlights the effectiveness of incorporating textual semantics for CVOC. Compared to our baseline obtaining 52.70\% and 58.30\% IoU for Ego$\rightarrow$Exo and Exo$\rightarrow$Ego, respectively, with an mIoU of 55.50\%, TeSCo achieves absolute gains of 7.70\%, 12.0\%, and 9.85\%, further demonstrating the benefit of text. In addition, compared to ObjectRelator~\cite{fu2025objectrelator}, which also uses textual information and achieves 35.30\% and 40.30\% IoU with an mIoU of 37.8\%, TeSCo yields substantial improvements of 25.10\%, 23.24\%, and 22.31\%, evidencing the advantage of our design in exploiting text for object correspondence.

\setlength{\columnsep}{8pt}%
\begin{wraptable}{r}{0.50\textwidth}
\setlength{\tabcolsep}{6pt}
	\centering
	\renewcommand{\arraystretch}{1.05}
    \caption{Results on the HANDAL-X test set.}
	\resizebox{0.485\textwidth}{!}{
    \begin{tabular}{rcc}
    \Xhline{1.2pt}
    \textbf{Method} & \textbf{Dataset} & \textbf{IoU ($\uparrow$)} \\
    \hline\hline
    \rowcolor[HTML]{f0f0f5}
    \multicolumn{3}{c}{\textbf{(a) zero-shot learning}}\\
    PSALM\textcolor{gray}{\scriptsize{[ECCV'24]}}~\cite{zhang2024psalm} & Ego-Exo4D & 39.90 \\
    ObjectRelator\textcolor{gray}{\scriptsize{[ICCV'25]}}~\cite{fu2025objectrelator} & Ego-Exo4D & 42.80 \\
    CCMP\textcolor{gray}{\scriptsize{[CVPR'26]}}~\cite{yan2026learning}  & Ego-Exo4D & \underline{78.80} \\
    V$^{2}$-SAM\textcolor{gray}{\scriptsize{[CVPR'26]}}~\cite{pan2026v}  & Ego-Exo4D & 77.20 \\
    \arrayrulecolor{gray!50}
    \hline
    Baseline & Ego-Exo4D & 75.26 \\
    \rowcolor[HTML]{E8FAF9}
    TeSCo (ours) & Ego-Exo4D & \textbf{85.10} \\
    \arrayrulecolor{black}\hline\hline
    \rowcolor[HTML]{f0f0f5}
    \multicolumn{3}{c}{\textbf{(b) retraining}}\\
    PSALM\textcolor{gray}{\scriptsize{[ECCV'24]}}~\cite{zhang2024psalm} & HANDAL-X & 83.40 \\
    ObjectRelator\textcolor{gray}{\scriptsize{[ICCV'25]}}~\cite{fu2025objectrelator} & HANDAL-X & 84.70 \\
    CCMP\textcolor{gray}{\scriptsize{[CVPR'26]}}~\cite{yan2026learning}  & HANDAL-X & \underline{85.00} \\
    \arrayrulecolor{gray!50}
    \hline
    Baseline & HANDAL-X & 83.94 \\
    \rowcolor[HTML]{E8FAF9}
    TeSCo (ours) & HANDAL-X & \textbf{90.00} \\
    \Xhline{1.2pt}
    \end{tabular}}
        \label{tab:handal_x_res}
 \vspace{1pt}
\end{wraptable}
\vspace{0.2em}
\textbf{Results on HANDAL-X.} We further evaluate TeSCo on HANDAL-X. Following~\cite{yan2026learning,fu2025objectrelator}, we conduct experiments in zero-shot learning and retraining. In the zero-shot learning setting, the methods are trained on Ego-Exo4D and tested on HANDAL-X; in the retraining setting, the methods are both trained and evaluated on HANDAL-X. Tab.~\ref{tab:handal_x_res} reports the results and comparison.

As shown in Tab.~\ref{tab:handal_x_res} (a), although TeSCo is not trained on HANDAL-X, it achieves the best performance with an IoU of 85.10\%, surpassing the second best CCMP~\cite{yan2026learning} with 78.80\% IoU by 6.30\% and the baseline with 75.26\% IoU by 9.84\%, demonstrating strong cross-view generalization ability. When training with HANDAL-X, as shown in Tab.~\ref{tab:handal_x_res} (b), the IoU score of TeSCo is further improved to 90.0\%, outperforming all other methods and its baseline. These results consistently demonstrate the effectiveness of TeSCo and the importance of textual information for robust cross-view correspondence.

\subsection{Ablation Study}

To better understand TeSCo, we conduct extensive ablation studies in Ego$\rightarrow$Exo on the test set of Ego-Exo4D. Our final configuration in each ablation is highlighted in light blue. 

\vspace{0.2em}
\textbf{Impact of TCF and ITFR.} TCF and ITFR are key components in TeSCo. To verify their effectiveness, we conduct ablation studies in Tab.~\ref{tab:tcf_itm}. Please note that ITFR cannot function alone in TeSCo. Since it is designed to adapt textual features for fusion, it must be used together with TCF. As shown in Tab.~\ref{tab:tcf_itm}, without TCF and ITFR, our baseline achieves an IoU score of 52.70\% (\ding{182}). When TCF is applied to incorporate textual semantics, the IoU score is significantly improved to 58.74\% with 6.04\% absolute gains (\ding{182} \emph{v.s.} \ding{183}). When using our ITFR to adjust the textual feature, the performance of TeSCo is further improved to 60.40\% (\ding{182} \emph{v.s.} \ding{184}), indicating that better textual semantics could benefit more cross-view correspondence and the necessity of ITFR. These results clearly validate the efficacy of TCF and ITFR in exploiting textual semantics for improving CVOC.

\vspace{0.2em}
\textbf{Comparison of TCF and other fusion strategies.} TCF is presented to incorporate textual semantics by fusing textual features into visual representations. We compare TCF with two alternative fusion methods, including addition-based fusion and cross-attention-based fusion. The former simply adds the textual and visual features, while the latter injects textual information into visual features using standard cross-attention~\cite{vaswani2017attention}. The results are reported in Tab.~\ref{tab:fusion}. As shown, addition-based fusion achieves an IoU score of 56.43\% (\ding{182}). When utilizing cross-attention for fusion, the IoU score is improved to 58.26\% (\ding{183}). When applying TCF in TeSCo, we achieve the best performance with the IoU score of 60.40\% (\ding{184}), demonstrating the superiority of our TCF.

\vspace{0.2em}
\textbf{Impact of design choice in TCF.} In TeSCo, TCF applies the text as a condition to generate two key parameters, including reweight parameter $\boldsymbol{\gamma}_i$ and shift parameter $\boldsymbol{\beta}_i$, to further enhance the multimodal feature (see Eq. (\ref{mod})). We conduct ablation studies on $\boldsymbol{\gamma}_i$ and $\boldsymbol{\beta}_i$ in Tab.~\ref{tab:tcf_design}. As shown, when removing $\boldsymbol{\gamma}_i$ and $\boldsymbol{\beta}_i$, TCF degenerates into the standard cross-attention and achieves an IoU score of 58.26\% (\ding{182}). When adopting $\boldsymbol{\gamma}_i$ alone to reweight the feature, the IoU score is improved to 59.73\% (\ding{182} \emph{v.s.} \ding{183}). When using $\boldsymbol{\beta}_i$ alone to further inject textual guidance, the IoU score is enhanced to 58.80\% (\ding{182} \emph{v.s.} \ding{184}). When applying both $\boldsymbol{\gamma}_i$ and $\boldsymbol{\beta}_i$, we achieve the best IoU score of 60.40\% (\ding{182} \emph{v.s.} \ding{185}). All these results evidence the effectiveness of our design choice using $\boldsymbol{\gamma}_i$ and $\boldsymbol{\beta}_i$ in TCF.

\vspace{0.2em}
\textbf{Impact of number iterations in ITFR.} ITFR aims to iteratively refine the text feature using search view information in a progressive way. We conduct an ablation on the number of iterations $K$ in ITFR. As shown in Tab.~\ref{tab:iteration}, when setting $K=1$, which means adjusting the textual once, the IoU score is 59.90\% (\ding{182}). When further refining the textual feature with another iteration, \ie, $K=2$, the IoU score is improved to 60.40\% (\ding{183}), indicating further interaction enables better search-view-aware textual semantics. However, when setting $K=3$, the IoU score is slightly decreased to 59.29\% (\ding{184}), probably caused by over-refinement of the textual feature. Therefore, we set $K=2$ in this work.

\begin{figure*}[!t]
    \centering
    \begin{minipage}[t]{0.49\textwidth}
        \centering
        \captionof{table}{Ablation on TCF and ITFR in TeSCo.}\vspace{-3pt}
        \footnotesize
        \renewcommand{\arraystretch}{1.05}
        \setlength{\tabcolsep}{4.5pt}
        \scalebox{1}{
    \begin{tabular}{cccccc}
    \Xhline{1.2pt}
    & \textbf{TCF}   & \textbf{ITFR}   & \textbf{IoU ($\uparrow$)}   & \textbf{Cont.A ($\uparrow$)} & \textbf{Loc.E ($\downarrow$)} \\
    \hline\hline
    \ding{182}    &   -    &   -    & 52.70 & 0.719 & 0.027 \\
    \ding{183}    & \checkmark     &   -    & 58.74 & 0.798 & 0.016 \\
    \rowcolor[HTML]{E8F3FF}
    \ding{184}    & \checkmark     & \checkmark     & 60.40 & 0.847 & 0.011 \\
    \Xhline{1.2pt}
    \end{tabular}}
    \label{tab:tcf_itm}
    \end{minipage}
    \hfill
    \begin{minipage}[t]{0.49\textwidth}
        \centering
        \captionof{table}{Ablation on different fusion strategies.}\vspace{-3pt}
        \footnotesize
        \renewcommand{\arraystretch}{1.05}
        \setlength{\tabcolsep}{4pt}
        \scalebox{1}{
        \begin{tabular}{clccc}
        \Xhline{1.2pt}
          &       & \textbf{IoU ($\uparrow$)}   & \textbf{Cont.A ($\uparrow$)} & \textbf{Loc.E ($\downarrow$)} \\
          \hline\hline
        \ding{182}     & Addition & 56.43 & 0.781 & 0.018 \\
        \ding{183}     & Cross-Attention & 58.26 & 0.792 & 0.016 \\
        \rowcolor[HTML]{E8F3FF}
        \ding{184}     & Our TCF & 60.40 & 0.847 & 0.011 \\
        \Xhline{1.2pt}
    \end{tabular}}
        \label{tab:fusion}
    \end{minipage}

    \vspace{0.4cm} 

    \begin{minipage}[t]{0.49\textwidth}
        \centering
        \captionof{table}{Ablation on design choice in TCF.}\vspace{-3pt}
        \footnotesize
        \renewcommand{\arraystretch}{1.05}
        \setlength{\tabcolsep}{6pt}
         \scalebox{1}{
    \begin{tabular}{cccccc}
    \Xhline{1.2pt}
          &   $\boldsymbol{\gamma}_i$   &  $\boldsymbol{\beta}_i$ & \textbf{IoU ($\uparrow$)}   & \textbf{Cont.A ($\uparrow$)} & \textbf{Loc.E ($\downarrow$)} \\
    \hline\hline
    \ding{182}     & -& -& 58.26  & 0.792 & 0.016 \\
    \ding{183}     & \checkmark & -& 59.73 & 0.827 & 0.013 \\
    \ding{184}     & -& \checkmark & 58.80 & 0.800 & 0.015 \\
    \rowcolor[HTML]{E8F3FF}
    \ding{185}     & \checkmark & \checkmark & 60.40 & 0.847 & 0.011 \\
    \Xhline{1.2pt}
    \end{tabular}}   
        \label{tab:tcf_design}
    \end{minipage}
    \hfill
    \begin{minipage}[t]{0.49\textwidth}
        \centering
        \captionof{table}{Ablation on the number of iterations for text adaptation in ITFR.}\vspace{-4pt}
        \footnotesize
        \renewcommand{\arraystretch}{1.035}
        \setlength{\tabcolsep}{2.8pt}
        \scalebox{1}{
    \begin{tabular}{ccccc}
    \Xhline{1.2pt}
         &   \textbf{Num. of Iterations}     & \textbf{IoU ($\uparrow$)}   & \textbf{Cont.A ($\uparrow$)} & \textbf{Loc.E ($\downarrow$)} \\
         \hline\hline
    \ding{182}   & $K=1$   & 59.90 & 0.833 & 0.012 \\
    \rowcolor[HTML]{E8F3FF}
    \ding{183}   & $K=2$   & 60.40 & 0.847 & 0.011 \\
    \ding{184}   & $K=3$   & 59.29 & 0.794 & 0.017 \\
    \Xhline{1.2pt}
    \end{tabular}}
        \label{tab:iteration}
    \end{minipage}
\end{figure*}

\vspace{-0.5em}

\setlength{\columnsep}{8pt}%
\setlength\intextsep{-0pt}
\begin{wraptable}{r}{0.50\textwidth}
\setlength{\tabcolsep}{3pt}
	\centering
	\renewcommand{\arraystretch}{1.05}
    \caption{Ablation on text injection location.}\vspace{-3pt}
	\resizebox{0.47\textwidth}{!}{
    \begin{tabular}{cccccc}
    \Xhline{1.2pt}
          & \textbf{\makecell[c]{Inj. to \\Query View}} & \textbf{\makecell[c]{Inj. to \\Mem.}} & \textbf{IoU ($\uparrow$)}   & \textbf{Cont.A ($\uparrow$)} & \textbf{Loc.E ($\downarrow$)} \\
    \hline\hline
    \ding{182}     & \checkmark     &  -     & 55.28 & 0.753 & 0.022 \\
    \ding{183}     &   -    & \checkmark     & 58.94 & 0.804 & 0.015 \\
    \rowcolor[HTML]{E8F3FF}
    \ding{184}     & \checkmark     & \checkmark     & 60.40 & 0.847 & 0.011 \\
    \Xhline{1.2pt}
    \end{tabular}}
    \label{tab:text_loc}
 \vspace{3pt}
\end{wraptable}
\vspace{0.2em}
\textbf{Impact on Text Injection Location.} In TeSCo, we inject the textual cues to both query view and temporal memory. We conduct an ablation on injecting text to different locations in TeSCo, and report the results in Tab.~\ref{tab:text_loc}. As shown in Tab.~\ref{tab:text_loc}, injecting text into query view only yields 55.28\% IoU score (\ding{182}), while injecting text into memory alone improves to 58.94\% score (\ding{183}). When injecting text to both query-view feature and temporal memory, we achieve the best IoU score of 60.40\% (\ding{184}). Therefore, in this work, we inject the textual cues into both query view and temporal memory features.

\setlength{\columnsep}{8pt}%
\setlength\intextsep{-0pt}
\begin{wraptable}{r}{0.50\textwidth}
\setlength{\tabcolsep}{3pt}
	\centering
	\renewcommand{\arraystretch}{1.05}
    \caption{Ablation on TeSCo with different text generation methods. ``b'': box; ``m'': mask.}\vspace{-3pt}
	\resizebox{0.485\textwidth}{!}{
    \begin{tabular}{ccccc}
    \Xhline{1.2pt}
          & \textbf{Text Generation} & \textbf{IoU ($\uparrow$)}   & \textbf{Cont.A ($\uparrow$)} & \textbf{Loc.E ($\downarrow$)} \\
    \hline\hline
    \ding{182}     & Qwen2.5-VL + b & 56.28 & 0.766 & 0.020 \\
    \ding{183}     & Qwen2.5-VL + m  & 59.75 & 0.828 & 0.013 \\
    \rowcolor[HTML]{E8F3FF}
    \ding{184}     & DAM + m & 60.40 & 0.847 & 0.011 \\
    \color{gray!60}{\ding{185}} & \color{gray!60}{Manual Text} & \color{gray!60}{61.20} & \color{gray!60}{0.830} & \color{gray!60}{0.010} \\
    \Xhline{1.2pt}
    \end{tabular}}
    \label{tab:text_gen}
 \vspace{3pt}
\end{wraptable}
\vspace{0.2em}
\textbf{Impact of Text Generation Methods.} We adopt DAM~\cite{lian2025describe} for text generation. To show its effectiveness, we experiment TeSCo with Qwen2.5-VL-7B~\cite{abs-2502-13923} using respectively box and mask for text generation. As in Tab.~\ref{tab:text_gen}, using DAM for text generation achieves better IoU score of 60.40\% (\ding{184}). This is probably because DAM is trained for multi-granular region captioning, making it better suited for generating text. We further compare DAM with manual text provided by human. For each video, a global textual description containing category, color, texture, and material is labeled. As shown, TeSCo with DAM achieves performance very close to that with manual text (\ding{184} \emph{v.s.} \ding{185}), evidencing again the effectiveness of DAM for TeSCo.

Due to limited space, please refer to our \textbf{\emph{appendix}} for more experiments and analysis.

\section{Conclusion}

This paper introduces TeSCo, that exploits rich textual semantics of the target to improve CVOC. Specifically, TeSCo first generates a textual description of the target from the given query view, and then integrates the textual information into visual features for segmentation. To realize this, we design the iterative textual feature refinement (ITFR) and text-conditioned cross-modal fusion (TCF) modules in TeSCo. The former aims to adjust the textual feature using search-view information, while the latter works to incorporate the textual cues into visual features with text-guided modulation. In experiments on two datasets, TeSCo achieves state-of-the-art results, validating its efficacy. 

\textbf{Limitation.} Despite achieving promising results, TeSCo has two main limitations. First, it relies on frame-level descriptions, which may be degraded under occlusion or motion blur. In the future, we will explore temporal information in videos to generate more stable target descriptions. Second, TeSCo introduces additional computational resource and cost from text generation and textual feature extraction. In the future, we will optimize our method for more resource-friendly and efficient CVOC. 

\section*{Acknowledgments and Disclosure of Funding} 
This work receives no external funding. The authors declare no competing interests.



{
\bibliographystyle{plainnat}
\bibliography{reference}
}







\newpage
\appendix

\section*{\textbf{\Large{Appendix}}}

To better understanding of this work, we offer additional details, analysis, and results as follow:

\begin{itemize}
   \item \textbf{A \;\; Examples of Generated Textual Description} \\
   In this section, we show several examples of the generated textual descriptions of target.
  
   \item \textbf{B \;\; Architectures of Our Baseline} \\
   In this section, we show the detailed architecture of the baseline.
   
   \item \textbf{C \;\; Performance on Different Scenarios}\\
   In this section, we show the results on different scenarios in Ego-Exo4D.

   \item \textbf{D \;\; More Ablation Studies}\\
   In this section, we present more ablation studies of our method.

   \item \textbf{E \;\; Additional Results and Discussions}\\
   In this section, we present additional results and discussions.
   
   \item \textbf{F \;\; Complexity Analysis} \\
   In this section, we provide analysis of the model complexity for our proposed method and its comparisons with other state-of-the-art approaches.

   \item \textbf{G \;\; Qualitative Analysis}\\
   This section displays qualitative results of our method on different benchmarks.

   \item \textbf{H \;\; Broader Impact}\\
   This section discusses the broader impact of our method.
   
\end{itemize}

\section{Examples of Generated Textual Description}

\begin{figure}[bh]
    \centering
    \includegraphics[width=0.9\linewidth]{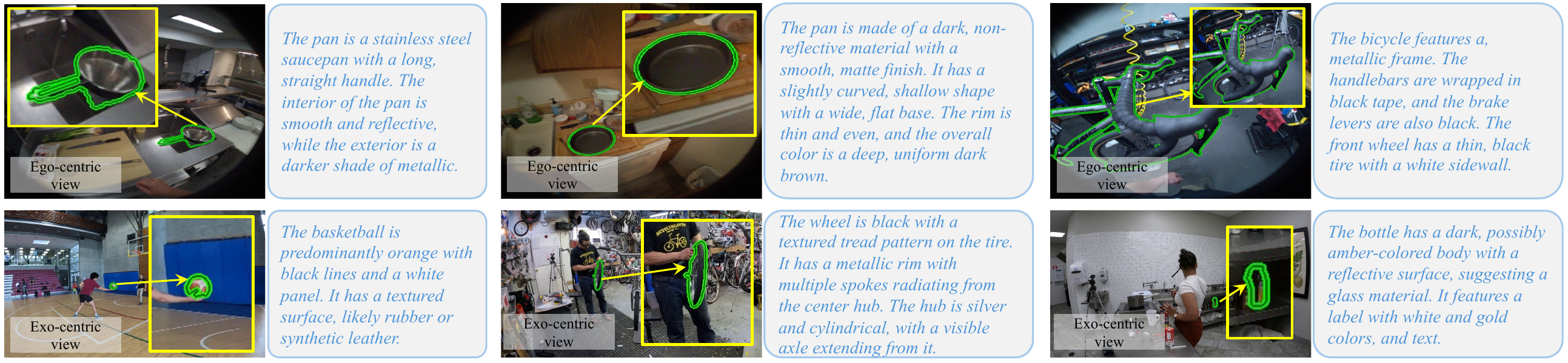}
    \caption{Examples of generated textual description.}
    \label{gen_text}
\end{figure}

Since text is not offered in the standard CVOC task, we apply an off-the-shelf vision-language model, DAM~\cite{lian2025describe}, to generate the textual description of the target for its excellent captioning ability. Fig.~\ref{gen_text} displays several examples of generated textual descriptions of the targets.

\vspace{0.5em}

\section{Architecture of Our Baseline}

\vspace{0.5em}

\begin{figure}[bh]
    \centering
    \includegraphics[width=0.8\linewidth]{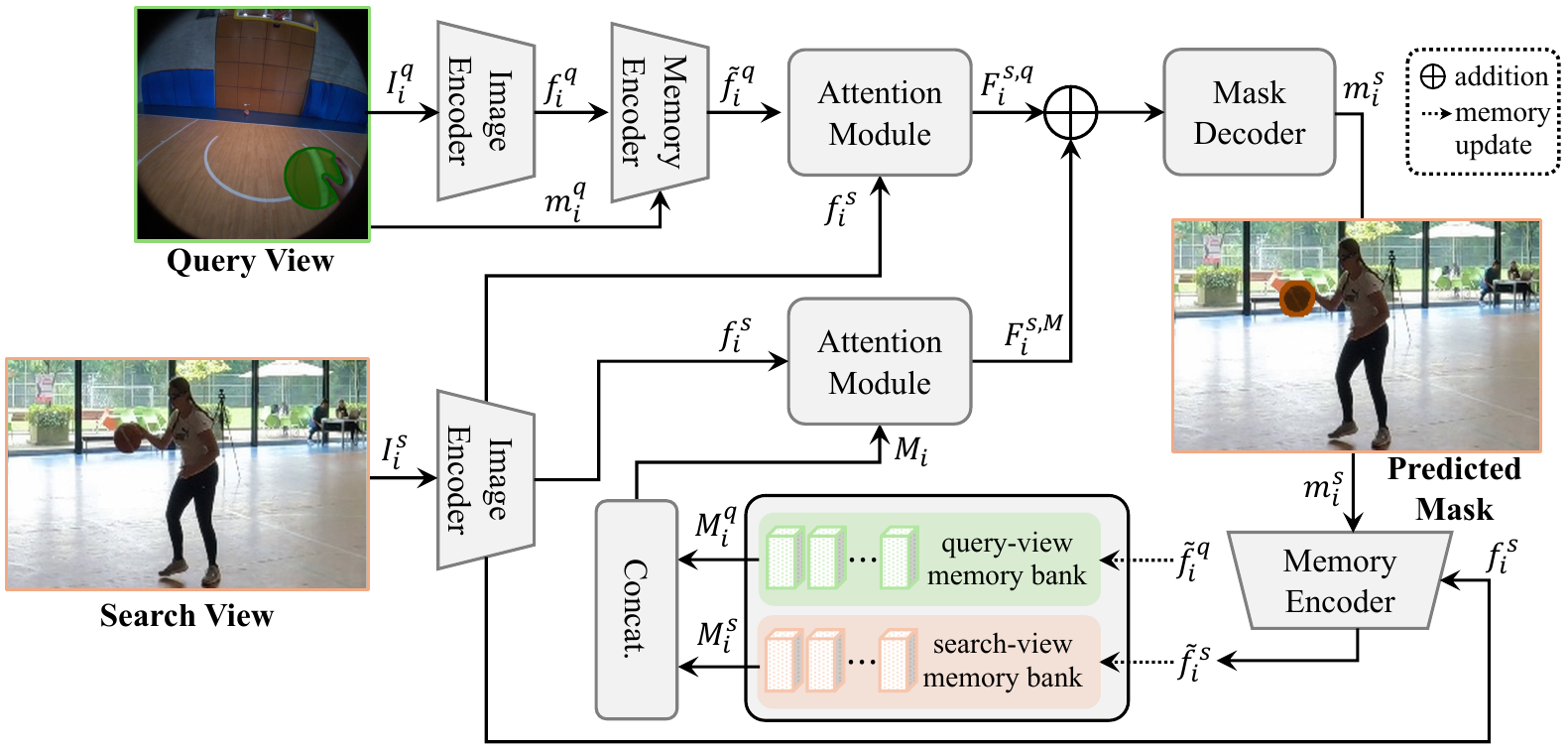}
    \caption{Illustration of the detailed architecture of our baseline.}
    \label{baseline}
\end{figure}

\vspace{0.5em}

This section illustrates the architecture of our baseline, which, based on the SAM 2~\cite{RaviGHHR0KRRGMP25}, exploits visual features, \ie, the query-view feature and temporal memory feature, for cross-view segmentation. Compared to TeSCo, the textual description generation, textual feature extraction, ITFR and TCF modules in the baseline are removed. Please refer to Fig.~\ref{baseline} for the framework of the baseline.

\section{Performance on Different Scenarios}

The Ego-Exo4D benchmark~\cite{grauman2024ego} comprises diverse scenarios, including \emph{Cooking}, \emph{Music}, \emph{Health}, \emph{Basketball}, \emph{Bike Repair}, and \emph{Soccer}. In order to further analyze our method, we show the performance on different scenarios. Tab.~\ref{tab:scenario_both} reports the results of TeSCo and our baseline. From Tab.~\ref{tab:scenario_both}, we can see that, TeSCo significantly the baseline model on all scenarios, showing the effectiveness of exploiting textual semantics in improving cross-view correspondence.

\begin{table}[!t]
	\centering
	\renewcommand{\arraystretch}{1.05}
	\setlength{\tabcolsep}{2.5pt}
	\footnotesize
	\caption{Performance on different scenarios in Ego-Exo4D.}\vspace{5pt}
	\label{tab:scenario_both}
	\resizebox{\linewidth}{!}{%
	\begin{tabular}{lcccccccccccc}
		\Xhline{1.2pt}
		\multirow{3}{*}{\textbf{Scenario}} & \multicolumn{6}{c}{\textbf{Ego$\rightarrow$Exo}} & \multicolumn{6}{c}{\textbf{Exo$\rightarrow$Ego}} \\
		\cmidrule(lr){2-7}\cmidrule(lr){8-13}
		& \multicolumn{3}{c}{\textbf{Baseline}} & \multicolumn{3}{c}{\textbf{TeSCo}} & \multicolumn{3}{c}{\textbf{Baseline}} & \multicolumn{3}{c}{\textbf{TeSCo}} \\
		\cmidrule(lr){2-4}\cmidrule(lr){5-7}\cmidrule(lr){8-10}\cmidrule(lr){11-13}
		& \textbf{IoU ($\uparrow$)} & \textbf{Cont.A ($\uparrow$)} & \textbf{Loc.E ($\downarrow$)} & \textbf{IoU ($\uparrow$)} & \textbf{Cont.A ($\uparrow$)} & \textbf{Loc.E ($\downarrow$)} & \textbf{IoU ($\uparrow$)} & \textbf{Cont.A ($\uparrow$)} & \textbf{Loc.E ($\downarrow$)} & \textbf{IoU ($\uparrow$)} & \textbf{Cont.A ($\uparrow$)} & \textbf{Loc.E ($\downarrow$)} \\
		\hline\hline
		Cooking & 44.93 & 0.685 & 0.027 & 57.29 & 0.829 & 0.131 & 53.76 & 0.665 & 0.005 & 68.54 & 0.805 & 0.025 \\
		Music & 72.72 & 0.789 & 0.020 & 88.93 & 0.940 & 0.010 & 60.96 & 0.708 & 0.034 & 74.55 & 0.843 & 0.017 \\
		Health & 48.07 & 0.669 & 0.019 & 61.93 & 0.847 & 0.008 & 57.52 & 0.678 & 0.041 & 74.11 & 0.859 & 0.018 \\
		Basketball & 70.35 & 0.875 & 0.019 & 79.32 & 0.956 & 0.007 & 80.08 & 0.889 & 0.010 & 90.29 & 0.971 & 0.004 \\
		Bike Repair & 26.39 & 0.427 & 0.068 & 53.08 & 0.774 & 0.021 & 30.75 & 0.403 & 0.122 & 61.85 & 0.732 & 0.038 \\
		Soccer & 53.24 & 0.672 & 0.026 & 72.06 & 0.911 & 0.008 & 58.47 & 0.686 & 0.039 & 71.38 & 0.853 & 0.018 \\
		\Xhline{1.2pt}
	\end{tabular}
	}
\end{table}

\section{More Ablation Studies}

In this section, we show more ablation studies of our method in Ego$\rightarrow$Exo on the test set of Ego-Exo4D. Our final configuration in each ablation is highlighted in light blue.

\setlength{\columnsep}{8pt}%
\setlength\intextsep{-0pt}
\begin{wraptable}{r}{0.50\textwidth}
\setlength{\tabcolsep}{3pt}
	\centering
	\renewcommand{\arraystretch}{1.05}
    \caption{Results with different textual description generation interval $r$.}\vspace{-3pt}
	\resizebox{0.485\textwidth}{!}{
    \begin{tabular}{ccccc}
		\Xhline{1.2pt}
		& \textbf{Generation Interval } & \textbf{IoU ($\uparrow$)} & \textbf{Cont.A ($\uparrow$)} & \textbf{Loc.E ($\downarrow$)}  \\
		\hline\hline
		\rowcolor[HTML]{E8F3FF}
		\ding{182} & $r=1$ & 60.40 & 0.847 & 0.011  \\
		\ding{183} & $r=5$ & 59.10 & 0.835 & 0.012  \\
		\ding{184} & $r=10$ & 59.00 & 0.835 & 0.013  \\
		\Xhline{1.2pt}
	\end{tabular}}
    \label{tab:gen_int}
 \vspace{3pt}
\end{wraptable}
\vspace{0.2em}
\textbf{Textual Description Generation Interval.} In this work, we generate the textual description from each query-view frame. This is because the target may reveal different visible attributes at different moments, and the frame-specific description can provide more suitable semantics for localization in the corresponding search-view frame. To validate this strategy, we vary the textual description generation interval $r$ among 1, 5, and 10 frames. An interval of 1 ($r=1$) corresponds to per-frame generation, whereas for intervals of 5 ($r=5$) and 10 ($r=10$), textual descriptions are generated every 5 and 10 frames, respectively, and the resulting textual features are reused for the subsequent intermediate frames until the next description is generated. Tab.~\ref{tab:gen_int} reports the results under the Ego$\rightarrow$Exo setting. As shown, generating textual descriptions at every frame achieves the best performance (\ding{182}), while increasing the interval to 5 or 10 frames leads to a gradual performance drop (\ding{183} and \ding{184}). This suggests that frame-specific textual descriptions benefits more the cross-view object correspondence.

\setlength{\columnsep}{8pt}%
\setlength\intextsep{-0pt}
\begin{wraptable}{r}{0.50\textwidth}
\setlength{\tabcolsep}{3pt}
	\centering
	\renewcommand{\arraystretch}{1.05}
    \caption{Impact of memory banks.}\vspace{-3pt}
	\resizebox{0.485\textwidth}{!}{
    \begin{tabular}{ccccc}
		\Xhline{1.2pt}
		& \textbf{Memory Bank} & \textbf{IoU ($\uparrow$)} & \textbf{Cont.A ($\uparrow$)} & \textbf{Loc.E ($\downarrow$)}  \\
		\hline\hline
		\ding{182} & query-view bank & 55.47 & 0.751& 0.022  \\
		\ding{183} & search-view bank & 58.78 & 0.800 & 0.016  \\
		\rowcolor[HTML]{E8F3FF}
		\ding{184} & using both banks & 60.40 & 0.847 & 0.017  \\
		\Xhline{1.2pt}
	\end{tabular}}
    \label{tab:mem}
 \vspace{3pt}
\end{wraptable}
\vspace{0.2em}
\textbf{Impact of Memory Banks.} In TeSCo, we employ both query-view and search-view memory banks to construct temporal memory features for robust cross-view object correspondence. To examine the impact of this dual-memory design, we conduct an ablation study by using only the query-view memory bank, only the search-view memory bank, and both memory banks. As shown in Tab.~\ref{tab:mem}, using both memory banks achieves the best performance, outperforming the variants that use only one of them. Specifically, using only the query-view memory bank yields an IoU of 55.47\% (\ding{182}), while using only the search-view memory bank improves the IoU to 58.78\% (\ding{183}). When both memory banks are applied, TeSCo further achieves the best IoU of 60.40\% (\ding{184}). These results indicate justifies the dual-memory design in TeSCo.

\section{Additional Results and Discussions}

\textbf{Targeted Stress Test.} To better understand our TeSCo, we perform textual and visual stress tests in Ego$\rightarrow$Exo on Ego4D. 

\setlength{\columnsep}{8pt}%
\setlength\intextsep{-0pt}
\begin{wraptable}{r}{0.45\textwidth}
\setlength{\tabcolsep}{5pt}
	\centering
	\renewcommand{\arraystretch}{1.05}
    \caption{Results on textual stress tests.}\vspace{-3pt}
	\resizebox{0.42\textwidth}{!}{
    \begin{tabular}{cccc}
		\Xhline{1.2pt}
		& \textbf{Method} & \textbf{Text Condition} & \textbf{IoU ($\uparrow$)}   \\
		\hline\hline
		\ding{182} & Baseline & No text & 52.70  \\
		\ding{183} & TeSCo & Original & 60.40   \\
		\ding{184} & TeSCo & Noisy & 59.22   \\
		\ding{185} & TeSCo & Wrong & 56.20   \\
		\ding{186} & TeSCo & Partial & 57.50   \\
		\Xhline{1.2pt}
	\end{tabular}}
    \label{tab:teststress}
 \vspace{3pt}
\end{wraptable}
\vspace{0.2em}
For textual stress tests, we construct three controlled textual perturbation settings: a) noisy captions, by randomly inserting one irrelevant semantic attribute into the description; b) wrong captions, by randomly replacing one key semantic attribute (\eg, category, color, or texture) with an incorrect one while preserving the remaining content; and c) partial captions, by randomly removing one key semantic attribute (\eg, category, color, or texture). Tab.~\ref{tab:teststress} reports the results on textual stress tests. From Tab.~\ref{tab:teststress}, we can observe that, although the performance decreases under challenging textual conditions, our method still outperforms the baseline, showing its robustness.

\setlength{\columnsep}{8pt}%
\setlength\intextsep{-0pt}
\begin{wraptable}{r}{0.45\textwidth}
\setlength{\tabcolsep}{3pt}
	\centering
	\renewcommand{\arraystretch}{1.05}
    \caption{Results on challenging visual subsets using IoU.}\vspace{-3pt}
	\resizebox{0.40\textwidth}{!}{
    \begin{tabular}{cccc}
		\Xhline{1.2pt}
		& \textbf{Evaluation Set} & \textbf{Baseline} & \textbf{TeSCo}   \\
		\hline\hline
		\ding{182} & Full test set & 52.70 & 60.40  \\
		\ding{183} & Occlusion subset	 & 48.25 & 58.50   \\
		\ding{184} & Distractor subset & 47.60 & 57.33   \\
		\Xhline{1.2pt}
	\end{tabular}}
    \label{tab:visualstress}
 \vspace{3pt}
\end{wraptable}
\vspace{0.2em}
For visual stress tests, we manually construct two subsets, Occlusion and Distractor subsets, from the full test set by inspecting all videos. Specifically, the Occlusion subset contains videos with occlusion, where the queried target is occluded by other objects. The Distractor subset contains videos with distractors which have similar appearance with the queried target. Tab.~\ref{tab:visualstress} shows the results challenging visual sub-
sets. We can see that, in challenging scenarios, both TeSCo and its baseline are degraded. However, TeSCo still significantly outperform its baseline, evidencing the effectiveness of textual semantics and the robustness of our method.

\setlength{\columnsep}{8pt}%
\setlength\intextsep{-0pt}
\begin{wraptable}{r}{0.45\textwidth}
\setlength{\tabcolsep}{3pt}
	\centering
	\renewcommand{\arraystretch}{1.05}
    \caption{Results of integrating our textual modules into LM-EEC.}\vspace{-3pt}
	\resizebox{0.40\textwidth}{!}{
    \begin{tabular}{ccc}
		\Xhline{1.2pt}
		& \textbf{Method} & \textbf{IoU}    \\
		\hline\hline
		\ding{182} & LM-EEC & 54.98  \\
		\ding{183} & LM-EEC + Our Textual Modules	 & 59.25    \\
		\Xhline{1.2pt}
	\end{tabular}}
    \label{tab:table_general}
 \vspace{3pt}
\end{wraptable}
\vspace{0.2em}
\textbf{Validation of Generality.} To validate the generality of our method, we conducting experiments by integrating the same textual modules of TeSCo, including, DAM-generated text, ITFR, and TCF, to LM-EEC~\cite{hu2025robust}. The results in Ego$\rightarrow$Exo on Ego4D is shown in Tab.~\ref{tab:table_general}. As shown, the performance improves from 54.98\% to 59.25\%, demonstrating that the gain mainly comes from textual semantics.

\textbf{Discussion of Textual Semantics in Challenging Scenarios.} In TeSCo, we leverage textual semantics of the target for cross-view segmentation. In challenging scenarios with the presence of similar objects, the discriminative power of text may decrease. However, in TeSCo, textual semantics serve as complementary guidance rather than the sole cue. The final correspondence is jointly determined by both the visual features and the textual semantics. Therefore, even when the textual descriptions become less discriminative, the visual information can provide fine-grained cues, and these two cues together can help localize the target object for robust segmentation.

\begin{figure}[bh]
    \centering
    \includegraphics[width=1\linewidth]{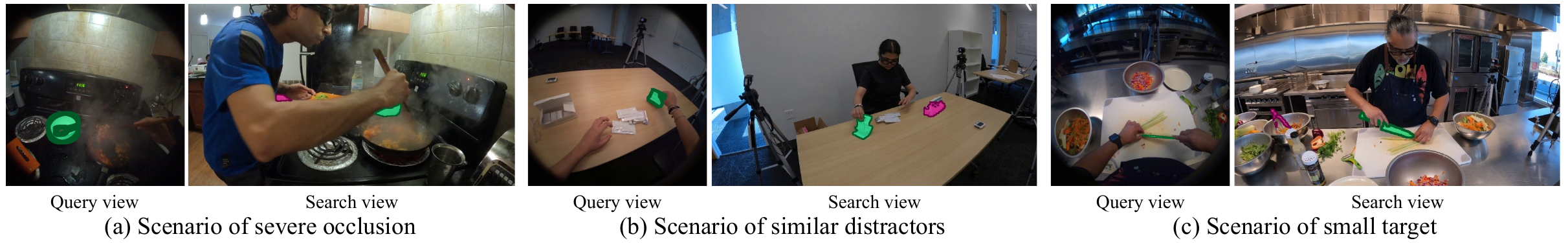}
    \caption{Representative failure cases of the proposed TeSCo under severe occlusion (see image (a)), similar distractors (see image (b)), and small target objects (see image (c)). The prediction results of TeSCo and groundtruth masks are highlighted in pink and green, respectively.}
    \label{failure_case}
\end{figure}

\textbf{Discussion on Failure Cases.} Although TeSCo achieves promising overall results, it may encounter failure cases in challenging scenarios, such as severe occlusion, visually similar distractors, and very small targets, as shown in Fig.~\ref{failure_case}. In these cases, reliable visual cues become insufficient, and the complementary textual semantics may also be less informative, making accurate cross-view correspondence more challenging and thus resulting in failures. To mitigate this, a future direction is to explore relational and temporal cues in videos to improve the discrimination power of the model.

\vspace{3mm}

\textbf{Discussion on More Implementation Details.} Here we provide more implementation details of our model, including (i) DAM checkpoint: the DAM checkpoint is ``nvidia/DAM-3B (the image variant)''; (ii) Text prompt: the DAM text prompt is ``use one to two sentence descriptions to describe what the object is (from appearance) from the given masked image region, including category, color, texture, and material if available''; (iii) Caption preprocessing: we do not perform preprocessing on the caption; (iv) RoBERTa pooling strategy: we apply the CLS-style pooling strategy; (v) Frozen and training components: during training, the DAM model and text encoder are frozen, and other components are trainable; (vi) Memory bank size: the memory bank size is 7; (vii) Loss weights: Following SAM 2, we use a focal mask loss (weight 20) that supervises pixel-wise mask classification, Dice loss (weight 1) that encourages mask overlap, IoU loss (weight 1) that improves mask localization quality, and occlusion binary cross-entropy loss (weight 1) that supervises occlusion prediction; (viii) Runtime computation: text generation is included in the runtime computation.

\textbf{Discussion on Uncertainty Estimates.} We report uncertainty estimates of our method on Ego4D. Specifically, our method achieves 60.32$\pm$0.19 mIoU in Ego$\rightarrow$Exo and 70.34$\pm$0.11 in Exo$\rightarrow$Ego (mean$\pm$standard deviation over three runs), indicating stable performance across different runs.

\textbf{Discussion on Consistency of Textual Semantics.} Our method is motivated by the observation that textual semantics provide complementary information to visual features, thereby improving target localization under cross-view appearance variations. Quantitatively, the performance improvements of TeSCo over the baseline, as reported in Tab.~\ref{tab:egoexo_res}, support the effectiveness of incorporating textual semantics. Qualitatively, the attention map visualizations in Fig.~\ref{attmap} show that textual semantics help the model attend more accurately to the target despite substantial appearance differences across views, suggesting textual semantics provide relatively consistent guidance across views.

\setlength{\columnsep}{8pt}%
\setlength\intextsep{-0pt}
\begin{wraptable}{r}{0.45\textwidth}
\setlength{\tabcolsep}{3pt}
	\centering
	\renewcommand{\arraystretch}{1.05}
    \caption{Results of integrating our textual modules into LM-EEC.}\vspace{-3pt}
	\resizebox{0.40\textwidth}{!}{
    \begin{tabular}{ccc}
		\Xhline{1.2pt}
		& \textbf{Method} & \textbf{IoU}    \\
		\hline\hline
		\ding{182} & Baseline & 52.70  \\
		\ding{183} & TeSCo with Category	 & 57.82    \\
        \ding{184} & TeSCo with Category and Color	 & 59.65    \\
        \ding{185} & TeSCo with Complete Description & 60.40    \\
		\Xhline{1.2pt}
	\end{tabular}}
    \label{tab:text_att}
 \vspace{3pt}
\end{wraptable}
\vspace{0.2em}
\textbf{Impact of Different Textual Attributes.} To study the impact of textual attributes in TeSCo, we conduct a systematic textual attribute ablation by progressively retaining more information from the descriptions: (i) Category, (ii) Category and Color, and (iii) Complete Description (\ie, all generated semantic attributes). Here, we select category and color because they are two of the most frequently occurring semantic attributes in the generated descriptions. The results are shown in Tab.~\ref{tab:text_att}. We can see that category information already provides substantial semantic cues for cross-view correspondence. Incorporating color further improves performance by enhancing object discrimination, while using the complete description achieves the best performance, demonstrating that richer semantic attributes provide complementary information beyond category and color.

\textbf{More Discussion with ObjectRelator~\cite{fu2025objectrelator}.} Our TeSCo share a similar spirit with ObjectRelator~\cite{fu2025objectrelator} in exploring textual cues for cross-view segmentation. Despite this, TeSCo  substantially differs from ObjectRelator, as discussed earlier in the paper. In experiments, compared to ObjectRelator, TeSCo shows more superior performance. Specifically on Ego4D, as shown in Tab.~\ref{tab:egoexo_res}, TeSCo with mIoU of 65.35\% surpasses ObjectRelator with 37.80\% by 27.55\%. On HANDAL-X, TeSCo also significantly outperforms ObjectRelator in Tab.~\ref{tab:handal_x_res}. These results show the advantages of our TeSCo in applying textual semantics for robust segmentation.

\setlength{\columnsep}{10pt}%
\setlength\intextsep{0pt}
\begin{wrapfigure}{r}{0.5\textwidth}
\centering
\includegraphics[width=0.5\textwidth]{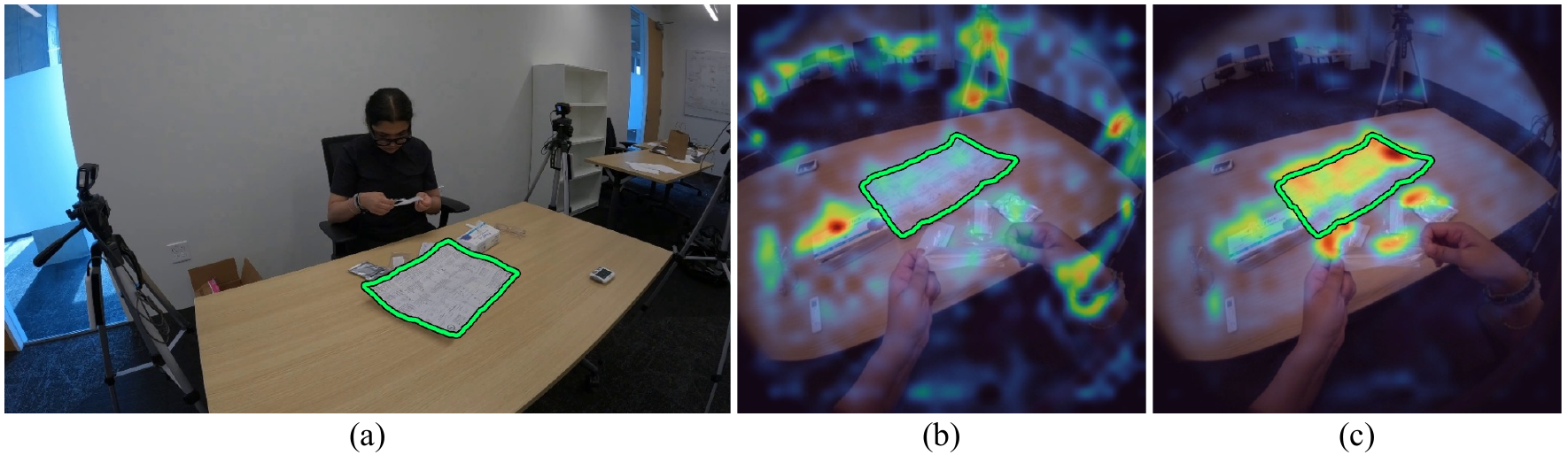}
\caption{(a) Query view with the target outlined in green. (b) Text-visual attention before ITFR. (c) Text-visual attention after ITFR.}
\label{fig:itfr_ref}
\end{wrapfigure}
\textbf{Effect of Text Refinement.} Fig.~\ref{fig:itfr_ref} visualizes the text-visual attention responses before and after ITFR refinement. Before refinement, the responses are scattered across the search view, with strong activations outside the target region. After refinement, the responses become more concentrated on the target object, indicating that ITFR effectively incorporates search-view visual context to enhance text-visual alignment.

\textbf{Discussion on (a) why the baseline model is strong and (b) why TeSCo improves substantially.} (a) The baseline is built upon the SAM 2, a general and powerful segmentation model. Since cross-view correspondence is essentially a segmentation task, it naturally benefits from the strong visual matching capability and temporal modeling ability of SAM2 for segmentation. As a result, the baseline, to some degree, can effectively establish robust visual correspondences across frames and viewpoints, making it a strong model for CVOC. (b) Although the baseline is strong, visual matching alone may still become ambiguous under large viewpoint and appearance variations. In these challenging scenarios, textual semantics provide relatively more stable cues that are less dependent on appearance and help preserve object identity across views. Building upon the strong baseline, TeSCo integrates the textual semantics into, rather than replacing, visual features, leading to stronger matching capability between different views and consistent improvements over the baseline model.

\section{Complexity Analysis}

In this work, we train the proposed TeSCo using 4 Nvidia A6000 (48GB) GPUs for around 36 hours. The total number of parameters is around 600M. In inference, the GPU memory requirement for running our TeSCo is 3.3GB and the inference speed is around 3 frames per second (\emph{fps}). A detailed inference-time breakdown include DAM with 272.7 ms on average, text encoder with 1 ms, ITFR with 0.5 ms, TCF with 0.6 ms, and SAM2 components (image encoder, memory attention, mask decoder, and memory encoder) with 24.4 ms. We can see that the runtime is dominated by DAM-based text generation, while the additional overhead of the text encoder, ITFR, and TCF is negligible. Compared with existing methods, including LM-EEC, V$^2$-SAM, CCMP, ObjectRelator, and O-MaMa, which run at 10, 2, 16, 1, and 4 FPS, respectively, TeSCo achieves a favorable balance between accuracy and inference speed.

\begin{figure}[!t]
    \centering
    \includegraphics[width=\linewidth]{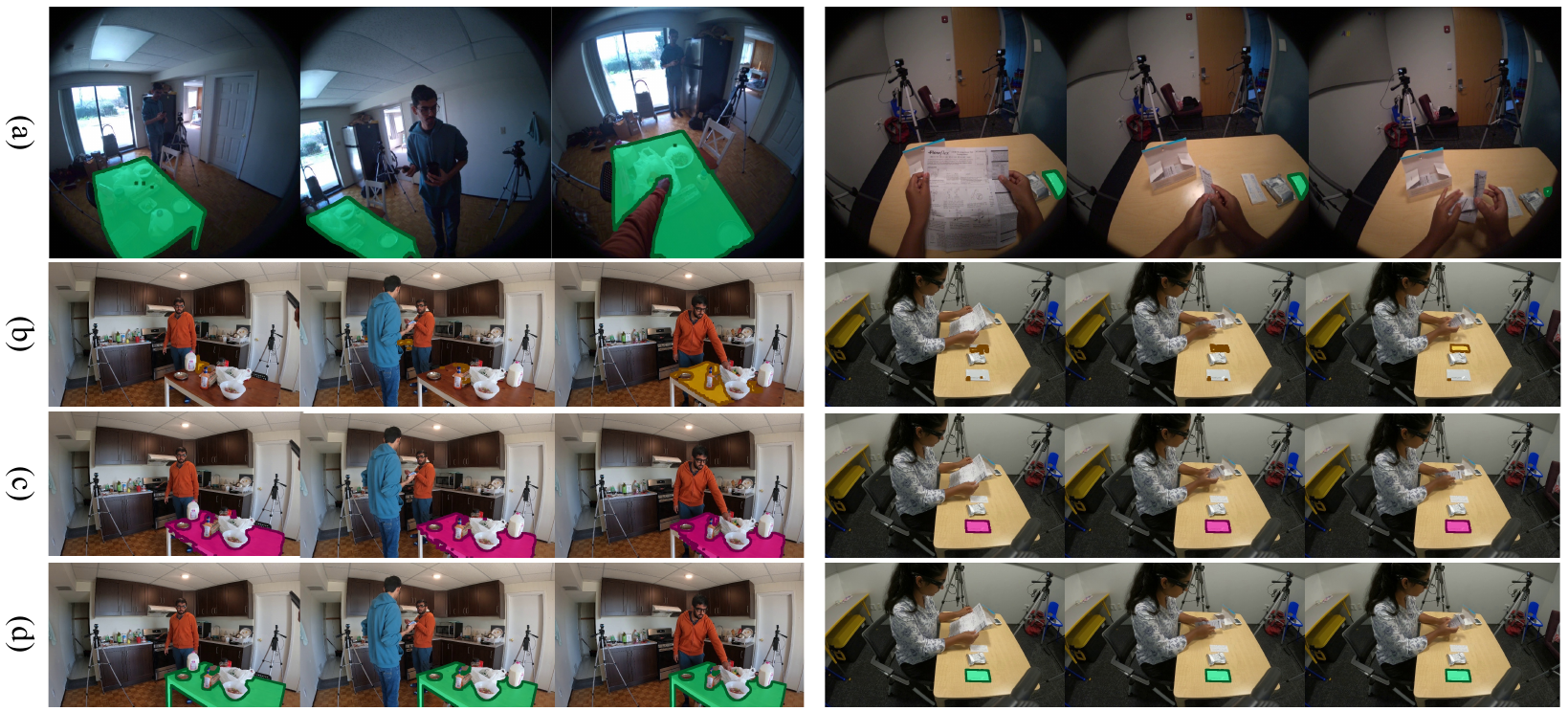}
    \caption{Qualitative results in Ego$\rightarrow$Exo. Image (a): query-view (Ego-centric); image (b): results of baseline; image (c): results of TeSCo; image (d): search-view (Exo-centric) and groundtruth mask.}
    \label{egoexo}
\end{figure}

\begin{figure}[!t]
    \centering
    \includegraphics[width=\linewidth]{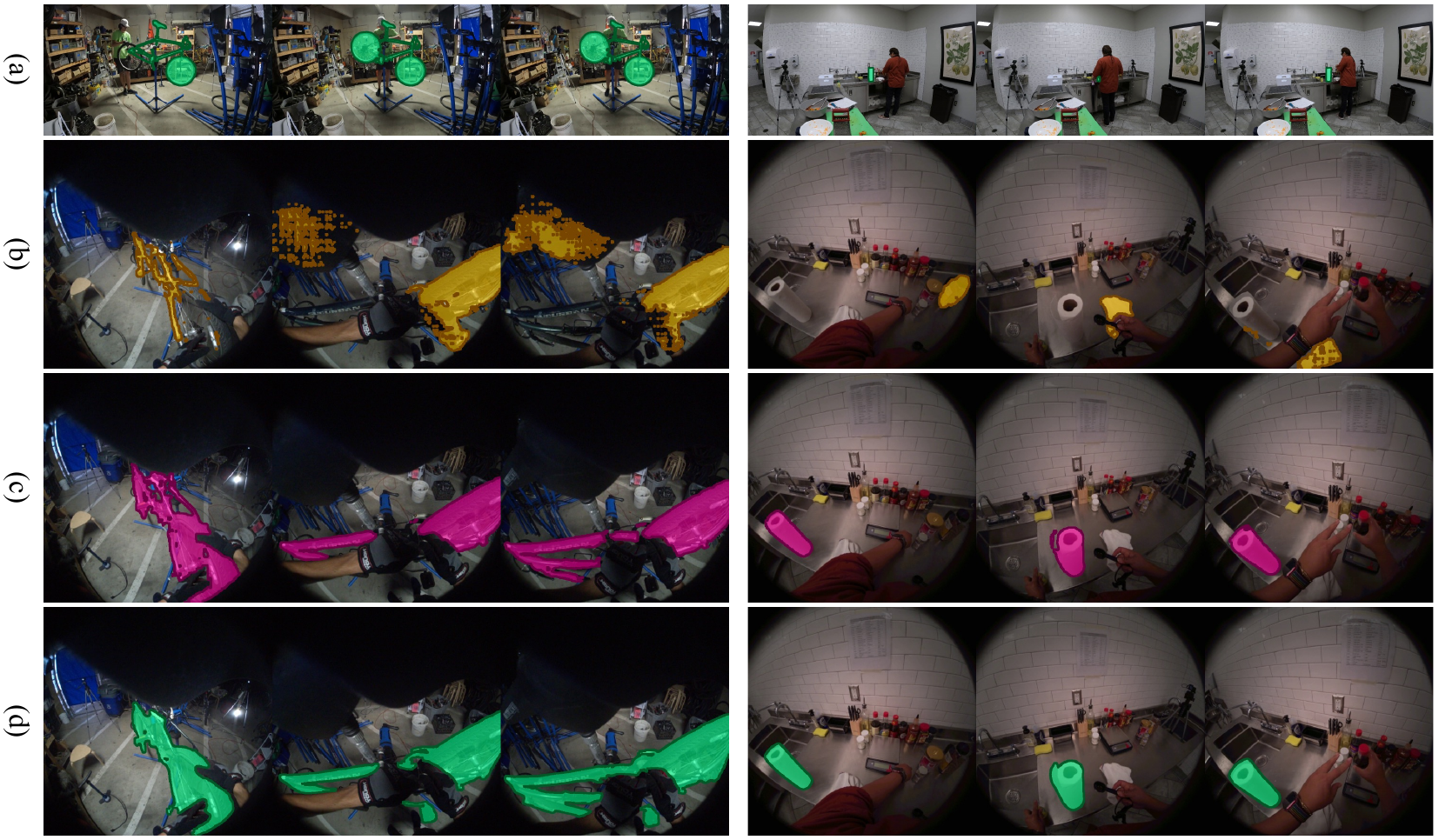}
    \caption{Qualitative results in Exo$\rightarrow$Ego. Image (a): query-view (Exo-centric); image (b): results of baseline; image (c): results of TeSCo; image (d): search-view (Ego-centric) and groundtruth mask.}
    \label{exoego}
\end{figure}

\begin{figure}[!t]
    \centering
    \includegraphics[width=\linewidth]{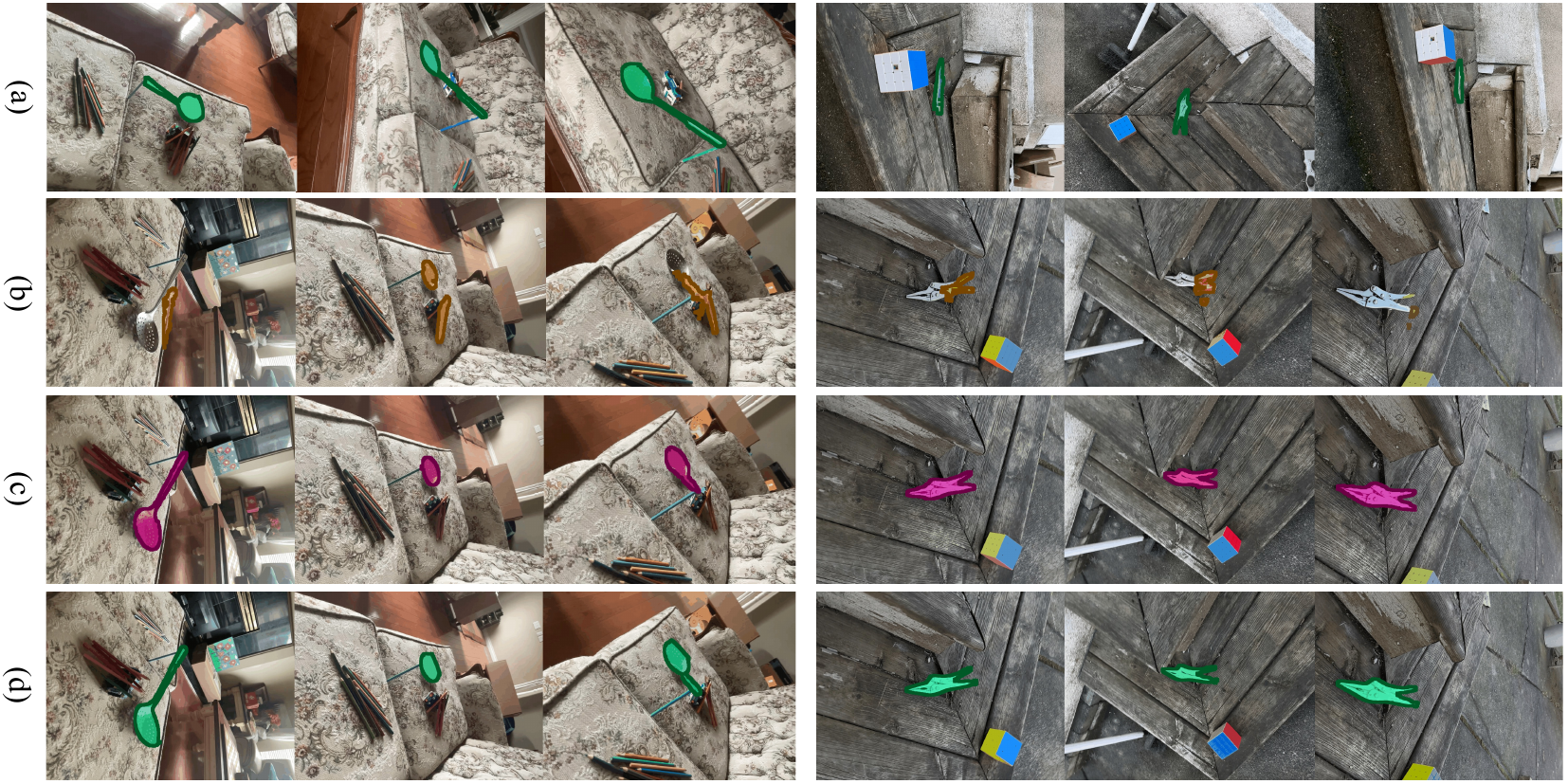}
    \caption{Qualitative results in HANDAL-X. Image (a): query-view; image (b): results of baseline; image (c): results of TeSCo; image (d): search-view and groundtruth mask.}
    \label{handalx}
\end{figure}

\section{Qualitative Analysis}

We further show the qualitative results of our method and comparison with the baseline model in Fig.~\ref{egoexo}, Fig.~\ref{exoego}, and Fig.~\ref{handalx}. Specifically, Fig.~\ref{egoexo} and Fig.~\ref{exoego} respectively display the results in Ego$\rightarrow$Exo and Exo$\rightarrow$Ego on Ego-Exo4D, and Fig.~\ref{handalx} shows the results on HANDAL-X. As shown, the baseline model struggles in challenging scenarios. However, our model, owing to textual semantics, is able to accurately segment the targets in cross-view videos.

\section{Broader Impact} 

Our TeSCo aims to establish cross-view object correspondence between different views and has a wide range of applications, such as intelligent robotics, augmented reality (AR), and assistive technologies. By enhancing the ability of AI systems to reason about objects from multiple viewpoints, TeSCo is able to contribute to developing more robust human-AI interaction systems.

It is worth noting that our work may raise potential ethical risks, particularly related to privacy. Specifically, the ability to associate different views could be misused for unauthorized tracking or monitoring of individuals beyond the intended scope of application. Therefore, our proposed method must be deployed in an ethical way.

\end{document}